\documentclass[UTF8]{article}
\usepackage{ijcai20}

\usepackage{times}
\usepackage{soul}
\usepackage{url}
\usepackage[hidelinks]{hyperref}
\usepackage[utf8]{inputenc}
\usepackage[small]{caption}
\usepackage{graphicx}
\usepackage{amsmath}
\usepackage{amsthm}
\usepackage{amsfonts}
\usepackage{booktabs}
\usepackage{multirow}
\usepackage[ruled]{algorithm2e}
\title{Large-scale Uncertainty Estimation and Its Application in\\ Revenue Forecast of SMEs}

\author{
Zebang Zhang, Kui Zhao, Kai Huang, Quanhui Jia, Yanming Fang, Quan Yu
\affiliations
Ant Financial Services Group\\
\emails
\{zebang.zhzb, zhaokui.zk, kevin.hk, quanhui.jia, yanming.fym, jingmin.yq\}@antfin.com
}

\begin{document}

\maketitle

\begin{abstract}
The economic and banking importance of the small and medium enterprise (SME) sector is well recognized in contemporary society. Business credit loans are very important for the operation of SMEs,  and the revenue is a key indicator of credit limit management. Therefore, it is very beneficial to construct a reliable revenue forecasting model. If the uncertainty of an enterprise's revenue forecasting can be estimated, a more proper credit limit can be granted. Natural gradient boosting approach, which estimates the uncertainty of prediction by a multi-parameter boosting algorithm based on the natural gradient. However, its original implementation is not easy to scale into big data scenarios, and computationally expensive compared to state-of-the-art tree-based models (such as XGBoost). In this paper, we propose a Scalable Natural Gradient Boosting Machines that is simple to implement, readily parallelizable, interpretable and yields high-quality predictive uncertainty estimates. According to the characteristics of revenue distribution, we derive an uncertainty quantification function. We demonstrate that our method can distinguish between samples that are accurate and inaccurate on revenue forecasting of SMEs. What's more, interpretability can be naturally obtained from the model, satisfying the financial needs.
\end{abstract}

\section{Introduction}

The economic and banking importance of the small and medium enterprise (SME) sector is well recognized in contemporary society ~\cite{biggs2002small}. Business loans are very important for the operation of SMEs. However, it is also acknowledged that these actors in the economy may be under-served, especially in terms of finance ~\cite{lloyd2006removing}. This has led to significant debate on the best methods to serve this sector. A substantial portion of the SME sector may not have the security required for conventional collateral based bank lending, nor high enough returns to attract formal venture capitalists and other risk investors. The effective management of lending to SMEs can contribute significantly to the overall growth and profitability of banks ~\cite{abbott2011management}. Banks have traditionally relied on a combination of documentary sources of information, interviews and visits, and the personal knowledge and expertise of managers in assessing the risk of business loans. But today, financial institutions have also begun to use big data and machine learning to manage credit risk for the credit loans ~\cite{khandani2010consumer}. Revenue is a key indicator of credit limit management. Therefore, it is very beneficial to construct an effective revenue forecasting model for credit limit management.

Forecasting the revenue of SMEs is a very challenging task. Traditional machine learning methods for financial regression tasks like revenue forecasting, such as Gradient Boosting Machines (GBMs) ~\cite{friedman2001greedy}, utilize the nonlinear transformation of decision tree to get more robust predictions. 
But for regression tasks, current popular models such as GBMs can only provide point estimates (forecast expectations or medians) and cannot quantify the predictive uncertainty. In financial tasks, it is crucial to estimate the uncertainty in forecast. The real revenue of SMEs is heteroscedastic distribution. The small enterprise with relatively unstable operating conditions have more large variance then medium enterprise with relatively stable operating conditions. A proper credit limit cannot be granted if the uncertainty of an enterprise's revenue forecasting cannot be estimated. This is especially the case when the predictions are directly related to automated decision making, as probabilistic uncertainty estimates are important in determining manual fall-back alternatives in the workflow ~\cite{kruchten2016machine}. In order to quantify the uncertainty, we need to upgrade from point estimation models to probabilistic prediction models. Probabilistic prediction, which is the approach where the model outputs a full probability distribution over the entire outcome space, is a natural way to quantify those uncertainties.

Bayesian method and non-Bayesian method are state-of-the-art methods in probabilistic uncertainty estimation. Bayesian methods naturally generate predictive uncertainty by integrating predictions over the posterior, but we are only interested in predictive uncertainty and do not focus upon the concrete procedure of generating uncertainty in predicting revenue of SMEs. In practice, Bayesian methods are often harder to implement and computationally slower to train compared to no-Bayesian method, such as Neural Network models and Bayesian Additive Regression Trees (BART) ~\cite{chipman2010bart}. Moreover, sampling-based Bayesian method generally requires good statistical expertise and thus leads to poor ease-of-use. Nature Gradient Boosting (NGBoost) ~\cite{duan2019ngboost} as the state-of-the-art algorithm of non-Bayesian method uses the natural gradient to address the challenge that simultaneous boosting of multiple parameters from the base learners. They demonstrate empirically that NGBoost performs competitively relative to other models in its predictive uncertainty estimates as well as on traditional metrics. They use decision tree from scikit-learn ~\cite{pedregosa2011scikit} as the base learner, which is a single machine algorithm supports the exact greedy splitting. NGBoost can only work on small data sets due to the single machine limit.

In this paper, We further derive the natural gradient to make it suitable for large-scale financial scenarios. We study the fisher information of normal distribution and find that the updating procedure of its natural gradient can be further optimized. For normal distribution, we propose a more efficient updating method for the natural gradient, which can dramatically improve computational efficiency. The base learner of SN-GBM is classification and regression trees(CART)~\cite{breiman2017classification}, which is the most popular algorithm for tree induction. Compared with NGBoost, SN-GBM adapts a more efficient distributed decision tree based on approximate algorithm as the tree-based learner, which can improve the computational efficiency and robustness. We derive an uncertainty quantification function to distinguish between samples that are accurate and inaccurate. In financial scenarios, interpretability is always demanded because of the transparency requirements of financial scenarios. So we provide two kinds of interpretability include uncertainty. Through the uncertainty interpretability, we can know the factors that cause the predictive uncertainty. In addition, we utilize the uncertainty outcome to optimize the procedure of solving regression problems, such as feature selection.

We summarize our contributions as follows:
\begin{enumerate}
\item We propose SN-GBM for large-scale uncertainty estimation in real industry and provide interpretability of the model.
\item We apply uncertainty estimation algorithm to revenue forecasting of SMEs for the first time.
\item We explore a range of uses of uncertainty estimation in regression tasks, which can bring a new modeling perspective.
\end{enumerate}
\section{Related Work}
\textbf{Sales Forecast.} Since there have been fewer published works about revenue forecast, we refer to some research about sales forecast. Sales often determine revenue. Sales forecast plays a prominent role in business strategy for generating revenue. Previous month sale is found to be more prominent parameters influencing the sales forecast in ~\cite{sharma2012sales}. Previous revenue is also an important factor in our revenue forecast. The most commonly used techniques for sales forecasting include statistically based approaches like time series, regression approaches and computational intelligence method like fuzzy back-propagation network (FBPN). ~\cite{chang2006fuzzy} and  ~\cite{sharma2012sales} both use FBPN for sales forecasting. FBPN algorithm performs more robust than traditional multiple linear regression algorithms in ~\cite{sharma2012sales}, which indicates nonlinear models are more appropriate for non-linear regression tasks such as sales forecast.

\textbf{Gradient Boosting Machines.} Gradient Boosting Machines ~\cite{friedman2001greedy} is a widely-used machine learning algorithm, due to its efficiency, accuracy, and interpretability. It has been shown to give state-of-the-art results in structured data (such as Kaggle Competitions). Popular scalable implementations of tree-boosting methods include ~\cite{chen2016xgboost} and ~\cite{ke2017lightgbm}. We are motivated in part by the empirical achievement of tree-based methods, although they only provide homoscedastic regression. One of the key problems in tree boosting is to find the best split feature value. ~\cite{chen2016xgboost} efficiently supports exact greedy for the single machine version, as well as approximate splitting algorithm. Also, we refer to some engineering optimizations in xgboost and lightgbm.

\textbf{Uncertainty Estimation.} Approaches to probabilistic forecasting can be broadly be distinguished as Bayesian or non-Bayesain. Bayesian approaches (which include a prior and a likelihood) that leverage decision trees for structured input data include ~\cite{chipman2010bart},~\cite{lakshminarayanan2016mondrian} and ~\cite{he2019xbart}. Bayesian NNs learn a distribution over weights to estimate predictive uncertainty ~\cite{lakshminarayanan2017simple}. Bayesian approaches cost expensive computational resource and are not easy to develop distributed algorithm. We are only interested in predictive uncertainty and do not pay attention to the concrete process of generating uncertainty. So bayesian approaches are not in our consideration. A non-Bayesian approach is similar to our work is ~\cite{duan2019ngboost} which takes a natural gradient method to solve the problem that multi-parameter boosting. Such a heteroskedastic approach to capturing uncertainty has also been called aleatoric uncertainty estimation ~\cite{kendall2017uncertainties}. As well as NGBoost, uncertainty that arises due to dataset shift or out-of-distribution inputs ~\cite{shiftcan} is not in the scope of our work.

\section{SN-GBM}
In the theory of algorithm, we mainly refer to the work of NGBoost. Firstly, We will clarify how NGBoost uses natural gradient to implement probabilistic prediction. Then we will demonstrate the improvements we have made on the basis of NGBoost, include more efficient updating method for the natural gradient. Traditional models can only output the interpretability of expectation. While SN-GBM can output two kinds of interpretability include uncertainty. Finally, we implement robust and interpretable Scalable Nature Gradient Boosting based on the decision tree from Spark, which is significantly faster than NGBoost.
\subsection{NGBoost}
The target of traditional regression prediction methods is to estimate $\mathbb{E}[y|\mathbf{x}]$. While the target of probabilistic forecast is to estimate $P_\theta(y|\mathbf{x})$, where $\mathbf{x}$ is a vector of observed features and $y$ is the prediction target, $\theta \in \mathbb{R}^p$ are parameters of target distribution. Take normal distribution for example, $\theta = [\mu, \sigma]$ (To be more specific, different $\mathbf{x}$ have different parameters $\mu, \sigma$, that is, $\theta = [\mu(\mathbf{x}), \sigma(\mathbf{x})]$).
\subsubsection{Proper Scoring Rules}
Fitting different targets need different loss functions. Probabilistic estimation requires "proper scoring rule" as optimization objective. A proper scoring rule $S$ takes as input a forecasted probability distribution $P$ and one observation $y$, and the true distribution of the outcomes gets the best score in expectation ~\cite{gneiting2007strictly}. In mathematical notation, a scoring rule is a proper scoring rule if and only if it satisfies
\begin{equation}
\mathbb{E}_{y\sim Q}[S(Q, y)] \le \mathbb{E}_{y\sim Q}[S(P, y)] \quad  \forall P, Q
\end{equation}
where $Q$ represents the true distribution of outcomes $y$, and $P$ is any distribution. When a proper scoring rule is used as loss functions during model training, the convergence direction of model is to output the calibration probability finally. In fact, maximum likelihood estimation (MLE), a method of estimating the parameters of a probability distribution, which satisfies above property. The difference from one distribution $Q$ to another $P$ is common KL divergence:

\begin{equation}
\begin{split}
D_S(Q||P)&=D_L(Q||P)\\
&=\mathbb{E}_{y\sim Q}\left[\log\frac{Q(y)}{P(y)} \right]
\end{split}
\end{equation}

It has a nice property that is invariant to the choice of parametrization ~\cite{dawid2014theory}. We will talk about importance of this property in later sections.

\subsubsection{Natural Gradient}
Gradient descent is the most commonly used method to optimize the objective function. The ordinary gradient of a scoring rule $S$
is the direction of steepest ascent (fastest increase in infinitesimally small steps). That is,
\begin{equation}
\nabla S(\theta, y)\propto \lim\limits_{\epsilon \to 0} \arg\max\limits_{d:||d||=\epsilon} S(\theta+d, y)
\end{equation}
However, ordinary gradient is not invariant to reparametrization. To be more specific, if we transform $\theta$ into $\psi = z(\theta)$, $P_{\theta + d \theta}(y)\ne P_{\psi + d \psi} (y)$. Therefore, different reparametrization approaches will affect the updating path of parameter. Again, we will talk about why we need invariant of the reparametrization.

The generalized natural gradient is the direction of steepest ascent in Riemannian space, which is invariant to parametrization, and is defined:
\begin{equation}
\tilde\nabla S(\theta, y)\propto \lim\limits_{\epsilon \to 0} \arg\max\limits_{d:D_S(P_\theta||P_{\theta+d})=\epsilon} S(\theta+d, y)
\end{equation}
when choosing MLE as proper scoring rule, we get:
\begin{equation}
\tilde\nabla S(\theta, y) = \tilde\nabla L(\theta, y)\propto \mathcal{L}_L(\theta)^{-1} \nabla L(\theta, y)
\end{equation}
where $\mathcal{L}_L(\theta)$ is the Fisher Information carried by an observation about $P_\theta$.
Note that a Fisher Information matrix is calculated for each sample.

\subsection{Scalable Natural Gradient Boosting}
In this section, we take the normal distribution as an example to demonstrate how to implement efficient large-scale distribution estimation.
\subsubsection{Simplify Computation}
The key of NGBoost is to calculate $\tilde\nabla L(\theta, y)$, which is equal to calculate $\mathcal{L}_L(\theta)^{-1} \nabla L(\theta, y)$. NGBoost calculates $\mathcal{L}_L(\theta)^{-1} \nabla L(\theta, y)$ by solving system of linear equations, whose time complexity is $O(N^3)$ (where $N=2$). This time complexity is relatively high for a single machine algorithm. Moreover, solving the system of linear equations is also not conducive to implementing distributed parallel algorithms. We find that on the premise of the normal distribution, a more direct method for calculating natural gradient can be derived.

The normal distribution is the most commonly used probability distribution. Many forecasting targets follow the normal distribution or can be transformed into a normal distribution (such as log-normal distribution). So we optimize the natural gradient calculation for the normal distribution. For normal distribution, the distribution parameters are $\theta = [\mu, \psi]$, where $\psi=\log(\sigma)$. By further derivation, we get:
\begin{equation}
\begin{split}
\nabla L(\theta, y)
=
\left[
 \begin{matrix}
   \frac{\mu - y}{\sigma^{2}} \\
1-\frac{(\mu-y)^2}{\sigma^{2}}
  \end{matrix}
  \right]
\end{split}
\end{equation}
Actually, the inverse of a fisher information matrix also can be derived simply. The fisher information of normal distribution is as follow:
\begin{equation}
\begin{split}
\mathcal{L}_L(\theta)=\mathbb{E}
 \left[
 \begin{matrix}
   \frac{1}{\sigma^{2}} & \frac{2 \left(- \mu + y\right)}{\sigma^{2}} \\
   \frac{2 \left(- \mu + y\right)}{\sigma^{2}} & \frac{2 \left(\mu - y\right)^{2}}{\sigma^{2}}
  \end{matrix}
  \right]
= \left[
 \begin{matrix}
   \frac{1}{\sigma^{2}} & 0 \\
   0 & 2
  \end{matrix}
  \right]
\end{split}
\end{equation}
Then, we can get:
\begin{equation}
\begin{split}
\mathcal{L}_L(\theta)^{-1}=
\left[
 \begin{matrix}
   \sigma^{2} & 0 \\
   0 & 0.5
  \end{matrix}
  \right]
\end{split}
\end{equation}
Finally, we derive the result of natural gradient:
\begin{equation}
\begin{split}
\tilde\nabla L(\theta, y)& \propto \mathcal{L}_L(\theta)^{-1} \nabla L(\theta, y) \\
& =
\left[
 \begin{matrix}
   \mu - y \\
0.5(1-\frac{(\mu-y)^2}{\sigma^{2}})
  \end{matrix}
  \right] \\
& =
\left[
 \begin{matrix}
   \mu - y \\
0.5(1-(\mu-y)^2 \exp(-2\psi))
  \end{matrix}
  \right]
\end{split}
\end{equation}

The second term is finally transformed into multiplication because the CPU of the computer calculates multiplication operations much faster than division. As we can see from the first term, NGBoost calculates the expectation $\mu$ in the same way as a normal gradient boosting machines that targets Mean squared error (MSE).
\subsubsection{Scalable Natural Gradient Boosting}
Gradient boosting is effectively a functional gradient descent algorithm. In order to fit multiple parameters of the distribution, we need multiple sets of trees, and each set of trees fits one parameter. Take normal distribution as an example, we use two sets of trees to fit $\mu$ and $\log(\sigma)$. Because the range of GBM output is $(-\infty, +\infty)$, but the range of $\sigma$ is $(0,+\infty)$. Reparameterizing $\sigma\in (0, +\infty)$ to $\psi = \log(\sigma)$, $\psi\in (-\infty, +\infty)$ is consistent with GBM output. This is one of the important reasons why natural gradient is needed: Natural gradient has the desirable property of being invariant to reparameterization. Another reason to use natural gradient is to enable using the same updating step size for two new trees when two sets of trees update at each stage. This because through the adjustment of $\mathcal{L}_L(\theta)^{-1}$, the gradient is scaled to the same scale whether it is between samples or parameters ("optimally pre-scaled").

Apache Spark is a popular open-source platform for large-scale data processing, which is specially well-suited for iterative machine learning tasks ~\cite{zaharia2010spark}. The MLlib ~\cite{meng2016mllib} ensemble decision trees for classification and regression problems. Decision trees use many state-of-the-art techniques from the PLANET project ~\cite{panda2009planet}, such as data-dependent feature discretization to reduce communication costs. Based on decision trees from Spark ML, we implement scalable natural gradient boosting machines, which is a tree and feature parallelization system. Since there is no dependency between two base learners at each iteration, two trees for two parameters can be constructed in parallel.

\begin{algorithm}
\caption{Scalable Natural Gradient Boosting for Normal Distribution}
\KwData{Dataset: $\mathcal{D}=\{x_i,y_i\}^{n}_{i=1}$}
\KwIn{Boosting iterations $M$, Learning rate $\eta$, Tree learner $f$, Normal distribution with parameters $\mu$ and $\psi = \log \sigma$, proper scoring rule $MLE$}
\KwOut{Scalings and tree learners $\{\rho ^{(m)},f^{(m)}\}^{M}_{m=1}$}
Initialize $\mu^{(0)} \leftarrow \frac{1}{n}\sum_{i=1}^{n}y_i$, $\psi^{(0)} \leftarrow \sqrt{\frac{1}{n-1}(y_i-\bar{y})^2}$ \\
\For{$m\leftarrow 1,...,M$}{
    \For{$i\leftarrow 1,...,n$}{
        $g(\mu_i)^{(m)} \leftarrow \mu_i^{(m-1)} - y_i$ \\
        $g(\psi_i)^{(m)} \leftarrow \frac{1-(\mu_i^{(m-1)}-y_i)^{2}\exp{(-2\psi_i^{(m-1)})}}{2}$
    }
    $f_{\mu}^{(m)} \leftarrow$ fit$(\{x_i,g(\mu)^{(m)}\}^{n}_{i=1})$ \\
    $f_{\psi}^{(m)} \leftarrow$ fit$(\{x_i,g(\psi)^{(m)}\}^{n}_{i=1})$
    $\rho ^{(m)} \leftarrow \arg\min_{\rho}\sum_{i=1}^{n}MLE(\mu_{i}^{m-1}-\rho\cdot f_{\mu}^{(m)}(x_i),\psi_{i}^{m-1}-\rho\cdot{f_{\psi}^{(m)}(x_i),y_i)}$ \\
    \For{$i\leftarrow 1,...,n$}{
        ${\mu_i^{(m)} \leftarrow \mu_i^{(m-1)}-\eta(\rho^{(m)}\cdot f_{\mu}^{(m)}(x_i))}$ \\
        ${\psi_i^{(m)} \leftarrow \psi_i^{(m-1)}-\eta(\rho^{(m)}\cdot f_{\psi}^{(m)}(x_i))}$
    }
}
\end{algorithm}

The overall training procedure is summarized in Algorithm 1. For normal distribution, $g(\mu)$ and $g(\psi)$ are the natural gradient of $\mu$ and $\psi$, respectively. In each iteration, two tree learners $f_{\mu}$ and $f_{\psi}$ will be constructed in parallel. The scaling factor $\rho$ is chosen to minimize MLE in the form of a line search. We multiply it by global update step $\eta$, then update the parameters to $\mu$ and $\psi$.
\subsubsection{Interpretability of Uncertainty}
For each sample, SN-GBM will output two prediction results, which are forecast expectation $\mu$ and variance $\sigma^2$. Theoretically, the smaller the variance, the narrower its distribution, and the more accurate the prediction. The heteroscedasticity of data often arises uncertainty. Heteroscedasticity often occurs when there is a large difference among the sizes of the observations. So we use variance to estimate the uncertainty of prediction results. In tree-based model, feature importance is often used as a factor in making decisions in interpreting models. SN-GBM is composed of two sets of trees, one is expectation set and another is variance set. We provide two approaches to getting the feature importance of variance:
\begin{enumerate}
\item Weight: The number of times a feature is used to split the data across variance trees.
\item Gain: The average gain of the feature when it is used in variance trees.
\end{enumerate}
By the feature importance of the variance, we can know which features affect the uncertainty of the prediction and the correlation score.

\section{Application in Revenue Forecast}
We propose an approach for quantifying the uncertainty of the forecasting target of the non-normal distribution of the original distribution. To provide a reliable and accurate prediction, we derive an uncertainty quantification function for revenue forecasting. Through the uncertainty quantification function, we can know the approximate probability of accurate predictions for each sample. In addition, we propose a bran-new feature selection based on the feature importance of variance, which can improve the precision of uncertainty quantification.
\subsection{Uncertainty Quantification}
The normal distribution is the most commonly used probability distribution. According to the central limit theorem, if an object is affected by multiple factors, no matter what the distribution of each factor is, the average of the results is a normal distribution. The normal distribution is symmetric, but many real-world distributions are asymmetric. Actually, if effects are independent but multiplicative rather than additive, the result may be approximately log-normal rather than normal. A Box-Cox transformation is a way to transform nor-normal dependent variables into a normal shape. One of the Box-Cox transformation is the log transformation. The real revenue distribution is close to a log-normal distribution. After log transformation, the revenue distribution has become a normal distribution, as shown in Figure 1.
\begin{figure}[ht]
\centering
\includegraphics[scale=0.23]{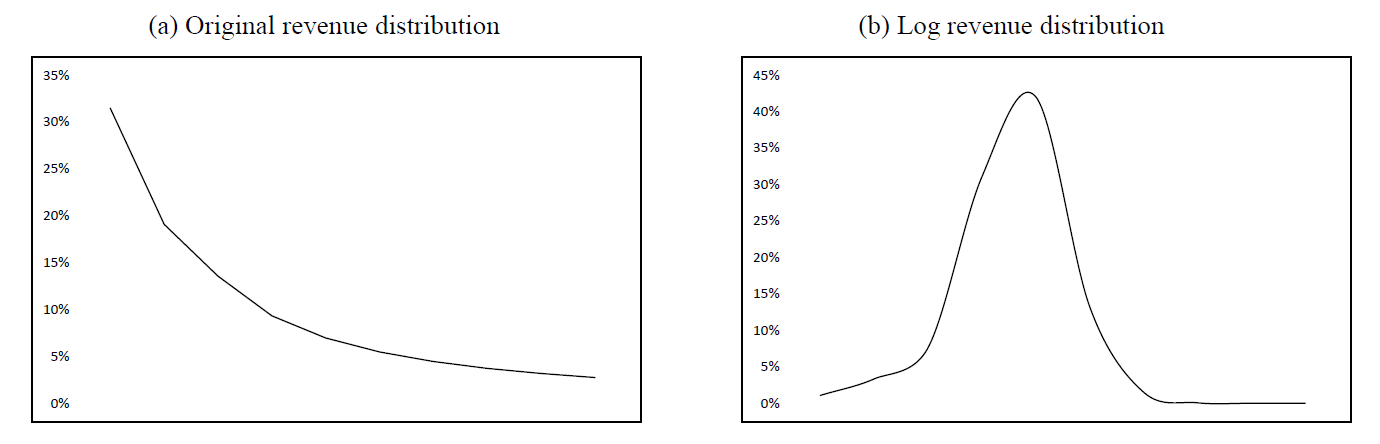}
\caption{(a) is the real revenue distribution, (b) is the log transformed revenue.}
\label{fig:label}
\end{figure}

In regression tasks, not only the error between the prediction and the observation is usually considered, but also the ratio between the error and the observation needs to be considered. In revenue forecasting, we train SN-GBM model to fit ln$(Y)$, where ln$(Y)\sim ~ \mathcal N(\mu,\sigma^2)$. So the $\mu$ and $\sigma$ of model output is the expectation and variance of ln$(Y)$. In fact, we need to estimate the uncertainty of the original revenue forecast by relative standard deviation, that is the ratio of the standard of $y$ to the expectation deviation of $y$. If the random variable ln$(Y)$ has a normal distribution, then the exponential function of ln$(Y)$, $Y$=exp(ln$Y$), has a log-normal distribution. $R$ notates relative standard deviation.

\begin{equation}
\begin{split}
R   &= \frac{\sqrt{[\exp(\sigma^2)-1]\exp(2\mu+\sigma^2)}}{\exp(\mu+\frac{\sigma^2}{2})} \\
    &= \sqrt{[\exp(\sigma^2)-1]}
\end{split}
\end{equation}

Because $\sqrt{[\exp(\sigma^2)-1]} \propto \sigma$, we can also use $\sigma$ to measure the relative standard deviation of $Y$.
\subsection{Feature Selection}
Data from many real-world applications can be high dimensional and features of such data are usually highly redundant. Identifying informative features has become an important step for data mining to not only circumvent the curse of dimensionality but to reduce the amount of data for processing. Feature Selection is the process where you automatically or manually select those features which contribute most to your prediction variable or output in which you are interested in. One of the commonly used approaches is to use the feature importance output by the tree-based model to filter features. The traditional tree-based models select the features that have a large contribution to the forecasting expectation based on the feature importance of the expectation. This method often ignores the correlation between features and uncertainty (here is variance). Based on the feature importance of the variance of SN-GBM, we can select features that are highly correlated with predictive uncertainty. Some features may have low expectation importance but have high variance importance. Such features may not improve the point estimation performance but may improve the accuracy of the distribution estimation. So in the future, we can combine the feature importance of expectation and variance to select features.

\section{Experiments}
\subsection{Datasets}
Our experiments use datasets from a large fintech services group. This group has served tens of millions of SMEs. One of the most significant scenarios is credit limit management. Our goal is to forecast the revenue of SMEs in the next six months. This is a time series regression task, so we mainly choose historical revenue and trade data of SMEs as the feature for constructing model. We extract twelve sub-datasets from January to December 2018 and five sub-datasets from January to May 2019. The first twelve months is the training set and the second five months is the test set. We extract 215 features related to the revenue for each enterprise. The size of training sample is 10 million.
\subsection{Evaluation Criteria}
Traditional evaluation metric is mean absolute percentage error (MAPE) of forecasted expectations (i.e. $\hat{\mathbb{E}}[y|x]$). It usually expresses accuracy as a percentage. Because the $\mu$ of SN-GBM output is the expectation of $\ln(y)$, so our MAPE formula is defined:
\begin{equation}
M=\frac{1}{n}\sum^{n}_{i=1}{|\frac{\exp(\mu)-y}{y}|}
\end{equation}
However, the MAPE does not capture predictive uncertainty. The quality of predictive uncertainty is captured in the average negative log-likelihood (NLL) as measured on the test set. NLL is calculated as follows:
\begin{equation}
NLL=-\frac{1}{n}\sum^{n}_{i=1}{\log \hat{P_\theta}(y_i|x_i)}
\end{equation}
where $P_\theta$ is the probability density function of the normal distribution $\hat{P_\theta}=\frac{1}{\sigma\sqrt{2\pi}}e^{-\frac{(y-\mu)^2}{2\sigma^2}}$.

In addition, in order to more intuitively apply the results of predictive uncertainty to credit limit management, we have added an evaluation metric \emph{ACCURACY}, which indicates the proportion of samples with prediction errors within 30\%. In the later section, we will briefly introduce how to utilize this metric for more refined credit limit management.
\subsection{Empirical Results}
\subsubsection{Results of Uncertainty Quantification}
We compare SN-GBM with several regression models commonly used in financial scenarios such as XGBoost ~\cite{chen2016xgboost} and GBDT ~\cite{friedman2001greedy}. For fair comparison, we set learning\_rate = 0.3, the number of iterations = 300, the depth of trees = 6 for all algorithms. Our experimental results show that SN-GBM is comparable to state-of-the-art tree-based models in the performance of point estimation, as shown in Table 1.

We sort the prediction results by the uncertainty quantification function $\sigma$, and then divide them into 10 buckets at equal samples. The uncertainty vs accuracy results is shown in Figure 2. The curve of accuracy and predictive uncertainty is monotonically decreasing. If the application demands an accuracy x\%, we can trust the model only in cases where the uncertainty is less than the corresponding threshold. For example, the \emph{ACCURACY} of top 50\% (from uncertainty level 1 to 5) samples is above 90\%. Because we have great confidence in prediction results of the top 50\% samples, for these enterprises we can directly use the predicted revenue as a reference factor for their credit limit. For other enterprises, we need to multiply the predicted revenue by a factor before using it.
\begin{table*}[t]
\centering
\setlength{\tabcolsep}{1mm}{
\begin{tabular}{|l|l|l|l|l|l|l|l|l|l|l|}
\hline
\multicolumn{1}{|c|}{\multirow{2}{*}{Algorithms}} & \multicolumn{2}{c|}{201901} & \multicolumn{2}{c|}{201902} & \multicolumn{2}{c|}{201903} & \multicolumn{2}{c|}{201904} & \multicolumn{2}{c|}{201905} \\ \cline{2-11}
\multicolumn{1}{|c|}{}                            & MAPE       & ACCURACY     & MAPE       & ACCURACY     & MAPE       & ACCURACY     & MAPE       & ACCURACY     & MAPE       & ACCURACY     \\ \hline
GBDT                                              & 0.310      & 0.687          & 0.305      & 0.712          & 0.294      & 0.736          & 0.320      & 0.724          & 0.293      & 0.742          \\ \hline
XGBoost                                           & \textbf{0.292}      & 0.694          & 0.303      & 0.711    & 0.290      & 0.697      & 0.314      & 0.720          & 0.277      & 0.754          \\ \hline
SN-GBM                                            & 0.301      & \textbf{0.735}          & \textbf{0.279}      & \textbf{0.746}          & \textbf{0.280}      & \textbf{0.746}          & \textbf{0.298}      & \textbf{0.712}          & \textbf{0.276}      & \textbf{0.755}          \\ \hline
\end{tabular}}
\caption{Comparison of performance of point estimation on revenue scenario. SN-GBM offers competitive performance of point estimation in terms of MAPE and ACCURACY.}
\end{table*}

\begin{figure}[ht]
\centering
\includegraphics[scale=0.2]{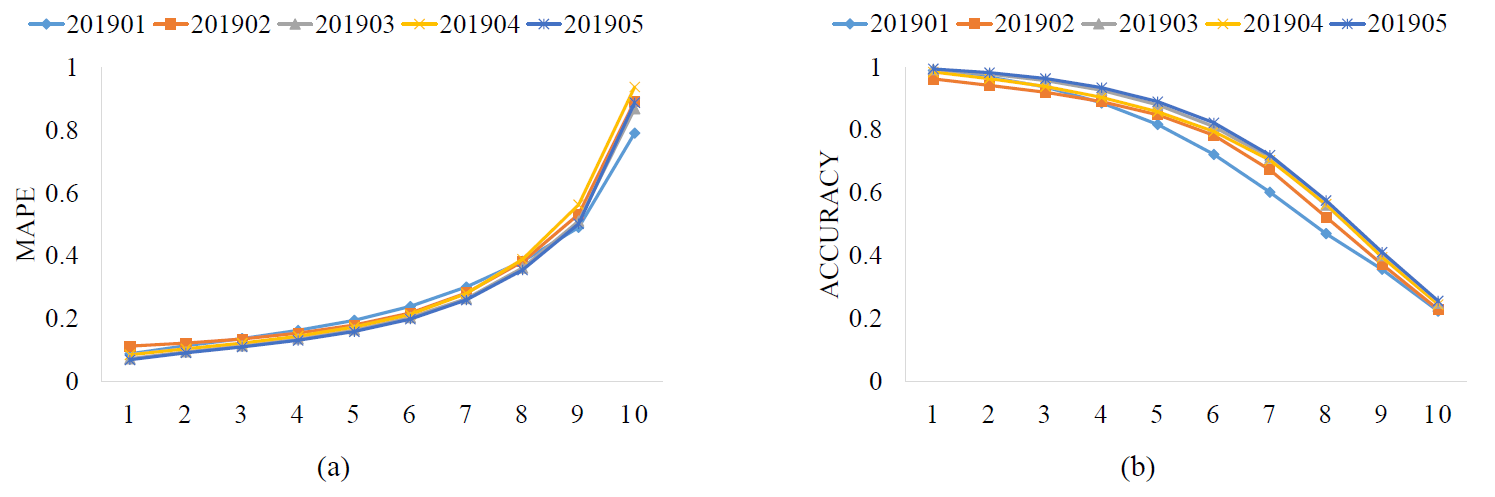}
\caption{Accuracy vs Uncertainty curves: The abscissa axis indicates the predictive uncertainty from 1 (low uncertainty) to 10 (high uncertainty).(a) MAPE performance in five test subsets. (b) ACCURACY performance in five test subsets.}
\label{fig:label}
\end{figure}
\subsubsection{Interpretability}
The interpretability of SN-GBM include expectation feature importance and variance feature importance. An example of the top 20 important features about expectation in revenue scenario is shown in Figure 3. From this figure, we observe that features with high expectation feature importance do not necessarily have high variance feature importance.
\begin{figure}[ht]
\centering
\includegraphics[scale=0.2]{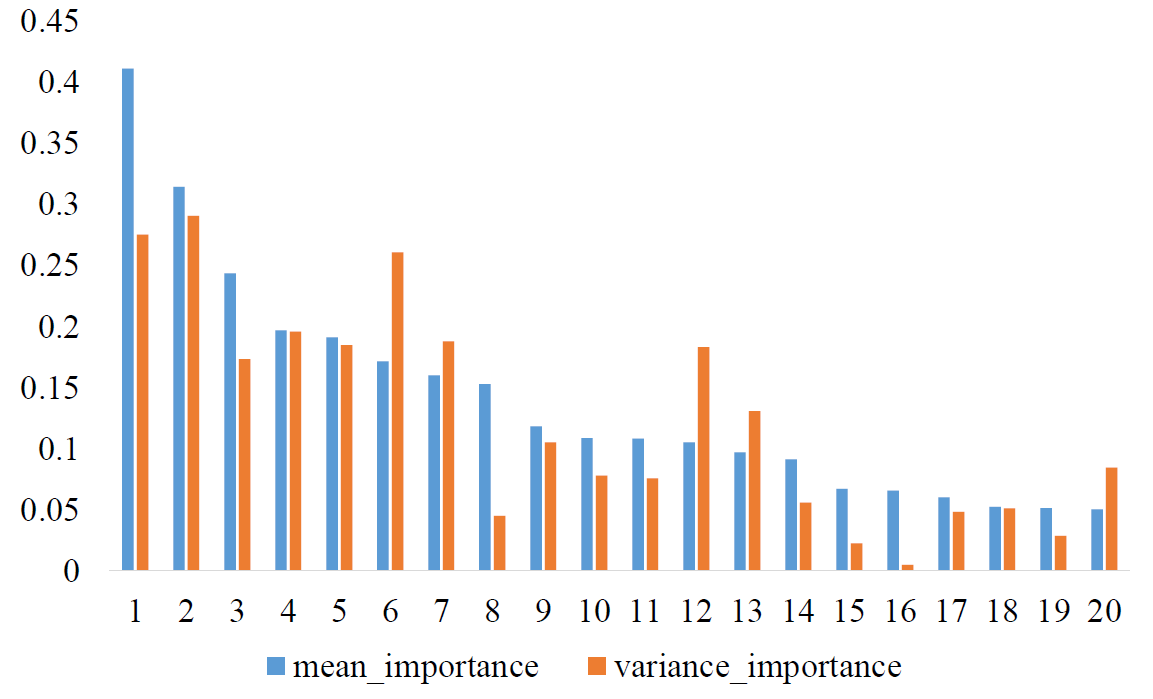}
\caption{Top 20 important features about expectation. The "mean\_importance" indicates the expectation feature importance, and the "variance\_importance" indicates the variance feature importance.}
\label{fig:label}
\end{figure}
\subsubsection{Feature Selection}
For the time series regression, the variance type features often better able to describe the predictive uncertainty. We append three time-series variance features to the 215 original features, which are the revenue variance in the past 3 months, the revenue variance in the past 6 months and the revenue variance in the past 12 months. We compare the performance of point estimation and distribution estimation of models with 215 features and 218 features (append three features about revenue variance), respectively. The results of point estimation are shown in Figure 4. After appending the features of revenue variance, the accuracy of model prediction has not brought a significant improvement. But from Figure 5 we can see, the accuracy of the distribution estimation is significantly improved, relative speaking.
\begin{figure}[ht]
\centering
\includegraphics[scale=0.2]{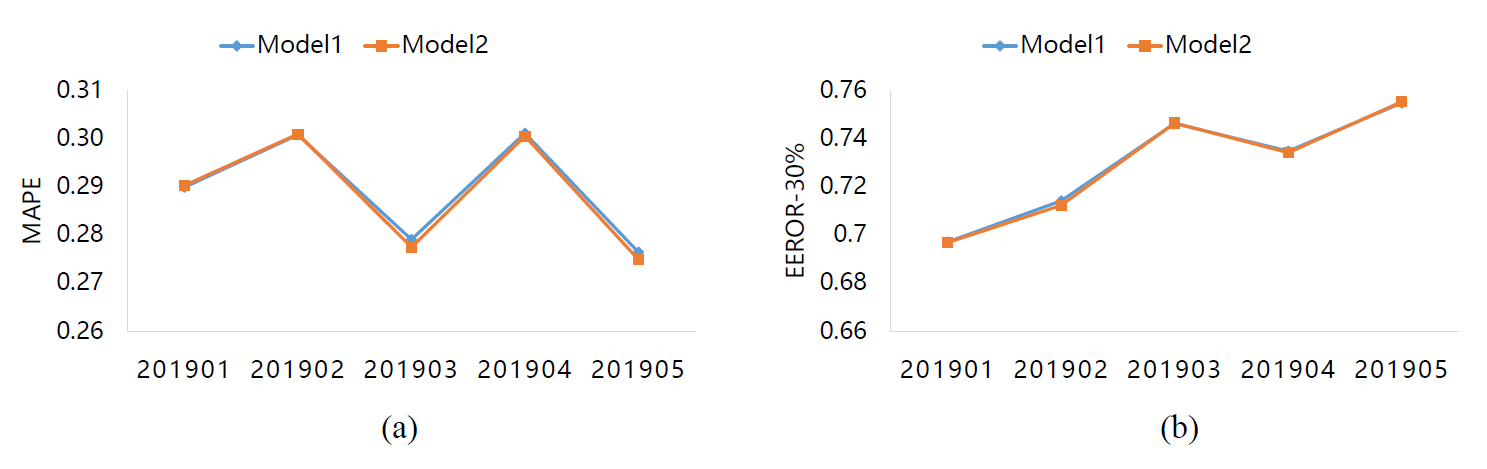}
\caption{The Comparison of point estimation performance. Model1: 215 features; Model2: 218 features.}
\label{fig:label}
\end{figure}

\begin{figure}[ht]
\centering
\includegraphics[scale=0.2]{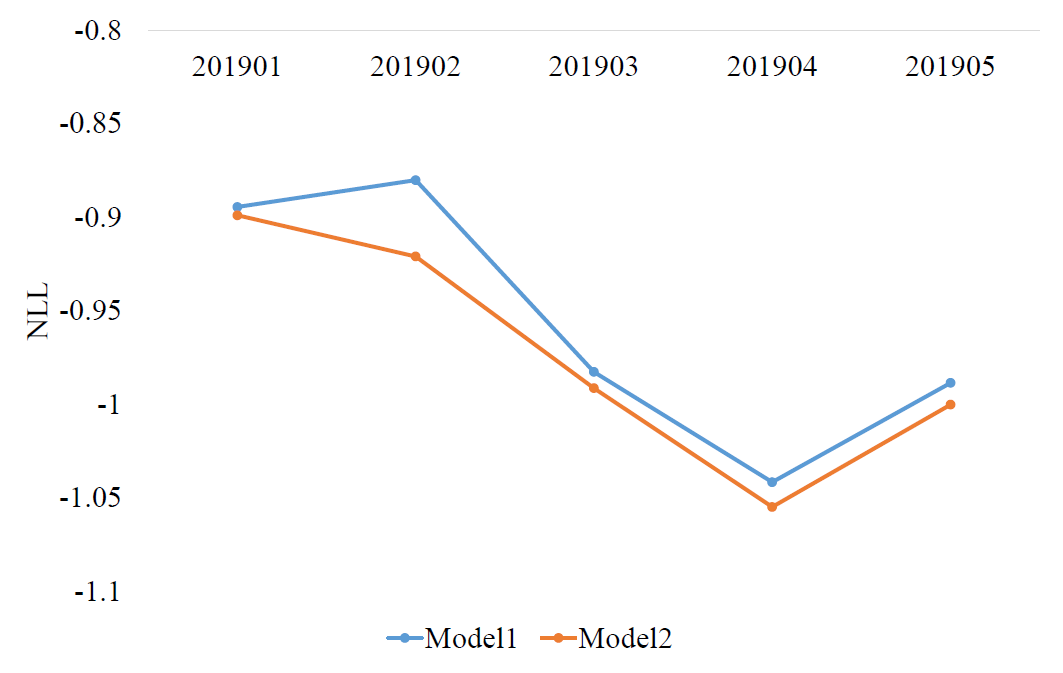}
\caption{The Comparison of distribution estimation performance. Model1: 215 features; Model2: 218 features.}
\label{fig:label}
\end{figure}

\section{Conclusion}
In this paper, we propose a large-scale uncertainty estimation approach named SN-GBM to predict the revenue of SMEs. The revenue distribution of SMEs is log-normal distribution. After log transformation, the revenue distribution is close to normal distribution. For normal distribution, we further derive the natural gradient to make it suitable for large-scale financial scenarios. We derive an uncertainty quantification function for the original distribution that is log-normal. Specially, we provide the interpretability for predictive uncertainty. Through the uncertainty interpretability, we can know the factors that cause the predictive uncertainty. Experimental results show that we can effectively distinguish between accurate and inaccurate samples on a large-scale real-world dataset, which is significantly beneficial for refined credit limit management. The features of the variance type can improve the accuracy of the distribution estimation. In the future, it is worth considering to retain features of the variance type when constructing regression models.

\clearpage
\bibliographystyle{named}
\bibliography{ijcai20}

\end{document}